\documentclass{article}

\PassOptionsToPackage{numbers, sort&compress}{natbib}
\usepackage[preprint]{neurips_2021}

\usepackage[utf8]{inputenc} % allow utf-8 input
\usepackage[T1]{fontenc}    % use 8-bit T1 fonts
\usepackage{hyperref}       % hyperlinks
\usepackage{url}            % simple URL typesetting
\usepackage{booktabs}       % professional-quality tables
\usepackage{amsfonts}       % blackboard math symbols
\usepackage{nicefrac}       % compact symbols for 1/2, etc.
\usepackage{microtype}      % microtypography
\usepackage{xcolor}         % colors

\usepackage{amsmath, subcaption, multiobjective}

\usepackage{tikz}
\usetikzlibrary{matrix, calc, chains, positioning, decorations.pathreplacing, shapes, arrows, fit, backgrounds}

\usepackage[colorinlistoftodos, prependcaption, textwidth=2cm, textsize=tiny]{todonotes}
\setuptodonotes{color=blue!29, bordercolor=blue!47}

\title{
Towards Optimally Weighted Physics-Informed Neural Networks in Ocean Modelling
}

\author{Taco de Wolff, Hugo Carrillo, Luis Martí \& Nayat Sánchez-Pi\\
Inria Chile Research Center\\
Av. Apoquindo 2827, piso 12, Las Condes, Santiago, Chile\\
\texttt{\{taco.dewolff,hugo.carrillo,luis.marti,nayat.sanchez-pi\}@inria.cl} \\
}

\begin{document}
\maketitle

\begin{abstract}
The carbon pump of the world's ocean plays a vital role in the biosphere and climate of the earth, urging improved understanding of the functions and influences of the ocean for climate change analyses. State-of-the-art techniques are required to develop models that can capture the complexity of ocean currents and temperature flows. This work explores the benefits of using physics-informed neural networks (PINNs) for solving partial differential equations related to ocean modeling; such as the Burgers, wave, and advection-diffusion equations. We explore the trade-offs of using data vs. physical models in PINNs for solving partial differential equations. PINNs account for the deviation from physical laws in order to improve learning and generalization. We observed how the relative weight between the data and physical model in the loss function influence training results, where small data sets benefit more from the added physics information.
%Additionally, we compare the variance of our results and analyze the implications of activation functions for training neural network derivatives.

% The carbon pump of the world's oceans plays a vital role in the biosphere and climate of the earth, urging improved understanding of the functions and influences of the oceans for climate change analyses. State-of-the-art techniques are required to develop models that can capture the complexity of ocean currents and temperature flows. We will explore the benefits of using physics-informed neural networks (PINNs) for solving partial differential equations related to ocean modeling; such as the wave, shallow water, and advection-diffusion equations. PINNs account for the deviation from physical laws in order to improve learning and generalization. However, in this work, we observe worse training and generalization results, possibly due the amount of data used in training. %contrary to recent publications.
\end{abstract}

%\listoftodos

\section{Introduction}

% - necesidad de las pinns para la modelación del océano
% - destacar novedad (no se ha aplicado en este dominio en particular) revisar en google scholar 
% - **clarificar este mensaje:** que esto es un primer paso (cuadradito de 1x1)
% - que cosa hemos visto que está "roto" (que punto queremos tratar y por qué?)
% - como lo vamos a tratar

The ocean plays a key role in the biosphere \citep{sanchez-pi-et-al-2020--neurips}. It regulates the carbon cycle by absorbing emitted CO\textsubscript{2} through the biological pump and dissipates a large part of the heat that is retained in the atmosphere by the remaining CO\textsubscript{2} and other greenhouse gases. The biological pump is driven by photosynthetic microalgae, herbivores, and decomposing bacteria. Understanding the drivers of micro- and macroorganisms in the ocean is of paramount importance to understand the functioning of ecosystems and the efficiency of the biological pump in sequestering carbon and thus abating climate change.

Similarly, there is also a clear scientific consensus about the effects of climate change on the ocean. Changes like the shift in temperature, an increase of acidification, deoxygenation of water masses, perturbations in nutrient availability, and biomass productivity to mention a few, have a dramatic impact on almost all forms of life in the ocean with further consequences on food security, ecosystem services, and the well-being of coastal communities and humankind in general.

Consequently, we end up with a cyclic dependency: to understand and create strategies to mitigate climate change it is of primordial importance to understand the ocean and to understand the ocean it is necessary to understand how climate change is impacting it. Therefore, this is not only an urgent but also a scientifically demanding task that must be addressed with a cohort approach involving different scientific areas: state-of-the-art artificial intelligence, machine learning, applied math, modelling, and simulation, and of course marine biology and oceanography.

That is why it can be hypothesized that it is necessary to develop new state-of-the-art artificial intelligence, machine learning, and mathematical modelling tools that will enable us to move forward with the understanding of our ocean and to understand, predict, and hopefully mitigate the consequences of climate change.

% AI and mathematical modelling tools to contribute to the understanding of the structure, functioning, and underlying eco-evolutionary mechanisms and dynamics of plankton in the global ocean. Methods like deep learning, causal and inference learning, sequential decision making, transfer learning, multi-criteria optimization are just a few that can be applied to these kinds of complex problems, allowing us to get reliable knowledge from the ocean and its interactions.

%The ocean is mostly a fluid in almost permanent movement -- subjected to currents, tides, waves and other interactions -- where different where organisms, atmosphere, and continents \taco{no entiendo}. That is why understanding and modelling the different aspects, global and local behavior, and interactions are essential.

The ocean is mostly a fluid in almost permanent movement -- subjected to currents, tides, waves and other interactions -- where different where organisms, atmosphere, and continents interact. Modelling oceanic fluid dynamics is essential but also very expensive in the computational sense. Mathematical models like the Navier-Stokes equations \citep{temam2001navier} can express most of the processes of interest \cite{mclean2012understanding}. However, the high computational cost involved in applying them with the correct precision level makes their application intractable. Machine learning (ML) and, in particular, neural networks are a competitive alternative. However, these approaches in general require a large amount of data to serve as examples for the learning process. In oceanographic research, obtaining data is challenging and expensive. The approach of hybrid models seems to be a good alternative to tackle this problem.

%In particular, we work on a first approximation to ocean models considering the wave equation, and we explore the \textit{physics-informed neural networks} (PINN) approach. We intend to show with numerical experiments that it is possible to recover the displacement field as well as the wave speed under the knowledge of those quantities in a small set of points in space and time, and also considering we know the wave equation holds. 

Physics-informed neural networks (PINNs) \citep{RAISSI2019686} are a hybrid approach that take into account a data-based neural network model and a physics-informed mechanistic model, which are two different paradigms. They offer a framework where existing knowledge about a physical phenomenon and empirical data gathered about it. This general concept has been previously explored and is known as data assimilation \cite{doi:10.1080/16000870.2018.1445364}, but PINNs bring a novel and sound approach to consolidate the existing models and sampled data. This feature makes PINNs particularly appealing for the above-described problems and has lead to some preliminary studies \cite{tacohugo-2021:aimocc,bjorn-2021:aimocc}.

Furthermore, it could be argued that by incorporating a rule that promotes the consistency of the neural with \emph{a priori} existing knowledge, the internal representations created during the training phase could be easier to understand and interpret by domain experts and, therefore, become not only a modelling but a research tool with a more broad impact.

This paper deals with the problem of predicting values of physical laws at given points in space and time relying on PINNs with a focus on ocean modelling. Our main goal is provide a first attempt -to the best of our knowledge- at determining what is the importance and relevance of the physical component in PINNs in improving the predictive power of a neural network concerning the amount of data available. By answering this question we aim to find a procedure for efficient configuration of parameters with numerical experiments inferred from the characteristics of the problem and the data available.

%The hyperbolic nature of the wave equation is a useful start for investigating typical hyperbolic ocean modelling equations such as the shallow water equations. It is possible to reduce the shallow water equations to the wave equation under certain suitable conditions, see for example \citep{stoker2011water}.
%On the other hand, the shallow water system is a basic model of fluids with relatively small vertical dimensions, such as for lake and coastal modelling. On a larger scale, we can think of the ocean being divided by layers of fluids \taco{is this true?} so that these shallow water equations could represent the top layer of the ocean. 
%Furthermore, coupling the shallow water and advection-diffusion equations could bring a simple toy model for the temperature in oceans. Hence our objective is to improve our understanding of ocean behaviour and its temperature flows to better model its interaction with climate change. \taco{we don't actually do this, why mention it?}

The contributions of this papers are: 
\begin{enumerate}
    \item to study the influence of the physics information and the scenarios where using it is more effective,
    \item a first step towards the automatic parameterization of PINNs using oceanographic modelling as test case, and
    \item we study the effects of other hyperparameters as the length and width of the neural network and activation functions on the optimal weight.
\end{enumerate}

The rest of this paper is organized as follows. In Section~\ref{sec:foundations} we present a brief rationale regarding the context of the paper, PINNs, modelling ocean fluids and the particular equations that we want to address (see~\ref{sec:ocean-pdes}).
% In Section~\ref{Model PDEs} we provide the values of the functions involved in partial differential equation (PDE) simulations and the essential aspects of PINN input generation. 
After that, in Section~\ref{sec:method}, we deal with the theoretical aspects of our work. Then, in Section~\ref{sec:experiments}, we describe and analyze the experimental studies involving PINNs for the estimation of the solution and parameters of the PDEs. Finally, in Section~\ref{sec:final} we put forward our conclusions of outline out future work. 

\section{Foundations}\label{sec:foundations}

High-dimensional partial differential equations (PDEs) \cite{rao2010introduction} are a common fixture in areas as diverse as physics, chemistry, engineering, finance, etc. Their enormous success has been hampered by its limited computational viability. Numerical methods for solving PDEs such as finite difference or finite element methods become infeasible in higher dimensions due to the explosion in the number of grid points and the demand for reduced time step sizes.

Addressing PDE scalability by approximating them by neural networks has been considered in various forms either by creating a training data set by directly sampling from an \emph{a priori} fixed mesh or by directly sampling the solution in a random manner (see \cite{Sirignano} for an overview). Consequently, a database of synthetic data coming from the numerical resolution of PDE-based models is generated on a broad range of scenarios. These data sets are then used to train and validate deep neural networks as in \citep{dupont-2019:augmented-odes, chen-2018:neural-odes}.

These approaches have the limitation of high computational costs for creating such data sets. To subvert the high computational costs, PINNs were recently proposed by \citep{RAISSI2019686} as a hybrid approach that considers a process where a source of physical knowledge in the form of PDEs is available. They specify the problem of training the neural networks as a multi-objective learning task, where we want to minimize the error with the data as well as the error with a physical law. Now we have three problems instead of one: minimizing the error with the data, minimizing the error with the physical law, and the combination of training for both objectives. \citep{RAISSI2019686} solves this by incorporating a physics loss term to the loss function, including a relative weight between both loss terms.

As it is a recent topic, there is not much literature about weighted PINNs. Having optimal weights for the loss functions in PINNs could help us to improve the performance of deep learning solvers for PDEs and so make the computations cheaper than regular PINNs. 
In \cite{van2020optimally}, the authors propose optimal weights for linear PDEs under the knowledge of boundary values and the differential equation. They use the structure and properties of the PDEs studied.

In \cite{xiang2021self}, the authors propose a self-adaptive loss balanced PINNs (lbPINNs) to solve the incompressible Navier-Stokes equation by an empirical search of the weights. In \cite{van2020physics}, the authors also chose empirically the weights to the loss functions in PINNs for study myocardial perfusion in MRI. On the other hand, in \cite{he2020physics} the authors evaluate the different results given by a few amount of weights in the loss functions for the advection-diffusion equation.

In this work, we study the influence of the weight on a validation quality metric, in particular the relative error of the neural network solution. In addition, we study the effects of other hyperparameters as the length and width of the neural network and activation functions on the optimal weight.

\subsection{Model equations related to the Ocean}\label{sec:ocean-pdes}

The ocean is characterized by its fluid nature. It is constituted mainly by water in different forms that is subjected to the effects of waves created by wind, currents, tides, among others. Therefore, in order to understand the ocean it is necessary to take into account this fundamental characteristic and to model it and the intervening phenomena. This gains particular relevance with the emergence of the climate change phenomenon and, therefore, we need trying to model, forecast and device policies to attempt to mitigate its effects.

Modelling oceanic fluid dynamics is essential but also very expensive in the computational sense. Mathematical models like the Navier-Stokes equations \citep{temam2001navier} can express most of the processes of interest. However, the high computational cost involved in applying them with the correct precision level makes their application intractable.

There are a number of equations that are of particular interest when modelling the ocean that are the focus of this work. In particular,
\begin{itemize}
    \item \emph{Burgers equation} \cite{BURGERS1948171}: appears often as a simplification to understand the main properties of the Navier-Stokes equations. It is a one-dimensional equation where the pressure is neglected but the effects of the nonlinear and viscous terms remain, hence as in the Navier-Stokes equations a Reynolds number can be defined \cite{Orlandi2000}. Hence, it is frequently used as a tool that is used to understand some of the inside behavior of the general problem.
    \item \emph{Wave equation} \citep{stoker2011water}: in this work we include the wave equation since, under certain suitable conditions, it can be seen as a reduced model of the shallow-water equation which models the water height behavior in coasts and channels. 
    \item \emph{Advection-diffusion equations} \cite{union2013advection}: models the behavior of temperature in a fluid, considering its velocity in the advective term and a diffusion phenomenon of the temperature. 
\end{itemize}

For the sake of briefness, a full description of these equations along with the details about how they are used in our experiments is given on Section~\ref{sec:experiments}.

In this work we study the behavior of the relative error (validation loss) with respect to the weight in the physical part, for Burgers, wave, and advection-diffusion equations.  In \cite{he2020physics} the authors study PINNs for advection-difussion equations, also considering different weights. On the other hand, it is very usual in the literature to find only the use of boundary and initial conditions as data, while a large amount of interior data is considered for data driven discovery of PDEs. 

In this work we consider a small amount of scattered data from the whole closed cylinder where the spatiotemporal PDEs are defined, which can simulate more likely the acquisition of sparse data in a huge domain. 

%In our work we present a large number of weights, and our setting for PINNs is not necessarily taking data from boundary or initial conditions, but from scattered interior points. \luis{revisar}

\section{Method}\label{sec:method}

As stated, our main goal is to understand the viability and particularities of the use of PINNs for oceanographic modelling. One particularity of this type of problem is the scarcity of data, specially if contrasted for the data-hungry needs of deep learning models that are currently the state of the art. That is why it is so important to understand the value that the physics information term can bring in order to ``fill the gaps'' where data is scarce or not available. 

We also hypothesize that, even in cases where there is a sufficient amount of data, adding a physics-informed loss function can potentially lead to a better interpretability of the trained model. This term can be seen as regularization term that would prompt the network to create representations that are consistent with the current knowledge of the phenomena. This, then should allow in the future to have an easier interpretation, assimilation, validation and acceptance by domain experts. 

PINNs are neural network models that are trained to obey laws in physics described by %(non-linear) 
partial differential equations (PDEs). They can be used used to solve supervised tasks in which we both minimize the error with respect to the data and to the physics law. We define the loss function as follows
\begin{equation}
\ell = (1-\lambda)\ell_\text{data} + \lambda \ell_\text{physics}
\end{equation}
where $\ell_\text{data}$ is the loss with respect to the data, $\ell_\text{physics}$ is the loss with respect to the PDEs of a physics law, and $\lambda \in [0,1]$ is the relative weight of the physics loss. If $f^j$ is the error of each physical condition, then using the mean-squared error for both losses we obtain%\luis{check this}\taco{done}
\begin{equation}\label{eqn:full-loss}
\ell = (1-\lambda)\sum^{N_u-1}_{i=0} \left(y_i - \hat{y}_i \right)^2 + \lambda \sum_j \sum^{N_f-1}_{i=0} (f^j_{i})^2\,.
\end{equation}

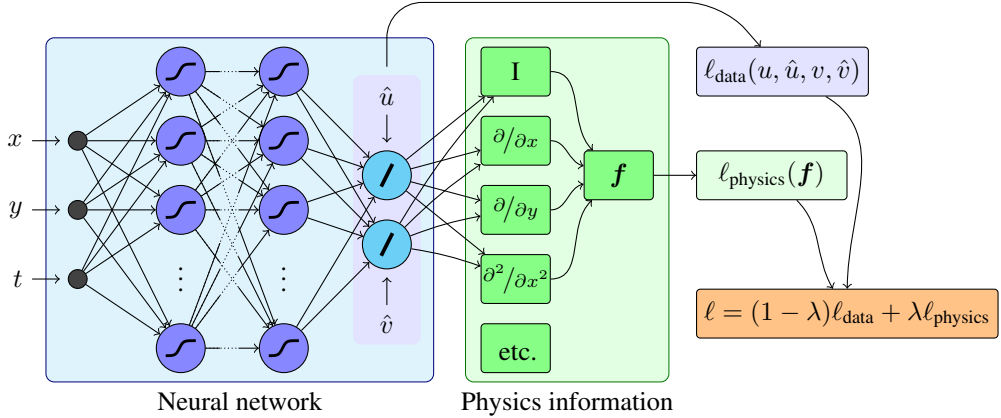
\begin{figure}[tb]
\centering
\begin{tikzpicture}
\def\layersep{1.37cm}
\def\neuronsep{0.92cm}

%\tikzset{node distance = \neuronsep and \layersep}

\newsavebox\logitbox
\savebox\logitbox{\draw[color=black,scale=0.2,line width=0.38mm] plot[mark=none, samples=25, domain=-8:8,] ({\x/8},{2/(2+exp(-\x))-0.5});;}

\newsavebox\linearbox
\savebox\linearbox{\draw[color=black,scale=0.15,line width=0.5mm] plot[mark=none,samples=2,domain=-3:3,] ({\x/6},{\x/3 - 0.0});;}

\tikzstyle{every pin edge}=[<-,shorten <=3pt]
\tikzstyle{neuron}=[circle, draw, color=black, fill=black!25, minimum size=0.65cm,inner sep=0pt]
\tikzstyle{input neuron}=[neuron, fill=black!74, minimum size=0.25cm];
\tikzstyle{output neuron}=[neuron, fill=cyan!50];
\tikzstyle{hidden neuron}=[neuron, fill=blue!47];
\tikzstyle{physics node}=[neuron, rectangle, fill=green!47, text width=0.92cm, text centered, rounded corners=0.05cm];
\tikzstyle{loss node}=[neuron, rectangle, fill=orange!47, text centered, minimum width=2cm, rounded corners=0.05cm, inner xsep=0.1cm];
\tikzstyle{ellipsis}=[neuron, draw=none, fill=none, align=center, text centered, text height=2.0ex];
\tikzstyle{annot} = [text width=4em, text centered]

\tikzstyle{neuralnet group} = [rectangle, line width=0.38pt, draw=blue!47!black, fill=cyan!11, text centered, rounded corners=0.11cm] % inner sep=1pt, text width=4.7cm, 
\tikzstyle{output group} = [rectangle, line width=0.38pt, draw=none, text centered, fill=blue!11, rounded corners=0.11cm]
\tikzstyle{physics group} = [rectangle, line width=0.38pt, draw=green!47!black, text centered, fill=green!11, rounded corners=0.11cm]

% Draw the input layer nodes
%\foreach \name / \y in {1,...,4}
% This is the same as writing \foreach \name / \y in {1/1,2/2,3/3,4/4}
%    \node[input neuron, pin=left:$x_{\y}$] (I-\name) at (0,-\y) {};

\node[input neuron, pin=left:$x$] (I-1) at (0,-\neuronsep) {};
\node[input neuron, pin=left:$y$] (I-2) at (0,-2*\neuronsep) {};
\node[input neuron, pin=left:$t$] (I-3) at (0,-3*\neuronsep) {};

% Draw the hidden layer nodes
\foreach \name / \y in {1,...,3}
    \path[yshift=\neuronsep] node[hidden neuron] (H1-\name) at (\layersep,-\neuronsep*\y) {\usebox\logitbox};
    
\node[ellipsis] (HE1) at (\layersep,-3*\neuronsep) {$\vdots$};
\node[hidden neuron] (H1-4) at (\layersep,-4*\neuronsep) {\usebox\logitbox};

% Draw the hidden layer nodes
\foreach \name / \y in {1,...,3}
    \path[yshift=\neuronsep] node[hidden neuron] (H2-\name) at (2*\layersep,-\neuronsep*\y) {\usebox\logitbox};

\node[ellipsis] (HE2) at (2*\layersep,-3*\neuronsep) {$\vdots$};
\node[hidden neuron] (H2-4) at (2*\layersep,-4*\neuronsep) {\usebox\logitbox};

% \node[input neuron] (I-0) [left=1.5cm of H-1,pin=north west:{\small${x^1_0=1}$}] {};

% Draw the output layer nodes
%\foreach \name / \y in {1,...,2}
%    \path[yshift=-0.5*\neuronsep, xshift=\layersep] 

\node[output neuron, pin={above:$\hat{u}$}] (O-1) at (3*\layersep, -1.5*\neuronsep) {\usebox\linearbox};
\node[output neuron, pin={below:$\hat{v}$}] (O-2) at (3*\layersep, -2.5*\neuronsep) {\usebox\linearbox};

\node[physics node] (F-1) at (4.25*\layersep, 0) {$\text{I}$};
\node[physics node] (F-2) at (4.25*\layersep, -\neuronsep) {$\nicefrac{\partial}{\partial x}$};
\node[physics node] (F-3) at (4.25*\layersep, -2*\neuronsep) {$\nicefrac{\partial}{\partial y}$};
\node[physics node] (F-4) at (4.25*\layersep, -3*\neuronsep) {$\nicefrac[\scriptsize]{\partial^2}{\partial x^2}$};
\node[physics node,text centered, text height=2.0ex] (F-etc) at (4.25*\layersep, -4*\neuronsep) {etc.};

\node[physics node] (F-big) at (5.25*\layersep, -1.5*\neuronsep) {$\vec{f}$};

\node[loss node, fill=blue!11, anchor=west] (L-nn) at (6*\layersep, 0) {$\ell_\text{data}(u, \hat{u}, v,\hat{v})$};
\node[loss node, fill=green!11, anchor=west] (L-physics) at (6*\layersep, -1.5*\neuronsep) {$\ell_\text{physics}(\vec{f})$};

\node[loss node, anchor=west] (L) at (6*\layersep, -3.5*\neuronsep) {$\ell = (1-\lambda)\ell_\text{data} + \lambda\ell_\text{physics}$};

% % Draw the output layer node
% %\node[output neuron, pin={[pin edge={->}]right:Output}, right of=H-3] (O) {};

% %\node[input neuron] (H-0) [above left=1cm of O-1,pin=north west:{\small${x^2_0=1}$}] {};

% % Connect every node in the input layer with every node in the
% % hidden layer.
\foreach \source in {1,...,3}
    \foreach \dest in {1,...,4}
        \path [->] (I-\source) edge (H1-\dest);
        % \path [->] (I-\source) edge node [sloped,near end,fill=white!98,inner sep=0pt] {\tiny $w_{\dest\source}$} (H-\dest);

\foreach \source in {1,...,4}
    \foreach \dest in {1,...,4}
        \draw[] (H1-\source) edge ($(H1-\source)!0.38!(H2-\dest)$) edge [densely dotted] ($(H1-\source)!0.56!(H2-\dest)$);
        
\foreach \source in {1,...,4}
    \foreach \dest in {1,...,4}        
        \path [->] ($(H1-\source)!0.56!(H2-\dest)$) edge (H2-\dest);
        
% \draw[] (A) edge ($(A)!0.45!(B)$) edge [dotted] ($(A)!0.55!(B)$);
\foreach \source in {1,...,4}
    \foreach \dest in {1,...,2}
        \path [->] (H2-\source) edge (O-\dest);
        % \path [->] (H-\source) edge node [sloped,near end,fill=white!98,inner sep=0pt] {\tiny $w_{\dest\source}$} (O-\dest);

% Connect every node in the hidden layer with the output layer
\foreach \source in {1,...,2}
    \foreach \dest in {1,...,4}
        \path [->] (O-\source) edge [bend right=5-5*\source] (F-\dest);

\foreach \source in {1,...,4}
    \path [->] (F-\source) edge [out=0, in=107+29*\source] (F-big);

% % Annotate the layers
% %\node[annot,above of=H-1] (hl) {Hidden layer};
% %\node[annot,above of=I-1] (il) {Input layer};
% %\node[annot,above of=O-1] (ol) {Output layer};

\begin{scope}[on background layer]
	\node [neuralnet group, inner xsep=0.29cm, fit=(I-1) (I-3) (H1-1) (H2-4) (O-2), label={below:Neural network}] (container) {};
\end{scope}

\begin{scope}[on background layer]
	\node [output group, inner ysep=1cm, fit=(O-1) (O-2)] (container-out) {};
\end{scope}

\begin{scope}[on background layer]
	\node [physics group, inner xsep=0.2cm, fit=(F-1) (F-etc) (F-big), label={below: Physics information}] (container-physics) {};
\end{scope}

\draw[->, shorten <=3pt, rounded corners=4.7mm] (container-out) -- (3*\layersep, \neuronsep) --  (6.25*\layersep, \neuronsep)  -- (L-nn);

% \path [->,[out=90, in=90]] (container-out) edge (3*\layersep, \neuronsep) (5.25*\layersep, \neuronsep) (L-nn);
\path [->] (F-big) edge [out=-0, in=180] (L-physics);
\path [->, pos=1.9] (L-nn) edge [out=-30, in=90] (L.north);
\path [->] (L-physics) edge [out=-45, in=120] (L);
\end{tikzpicture}
\caption{Schematic representation of a physics-informed neural network with inputs $x$, $y$, and $t$; outputs $\hat{u}$ and $\hat{v}$. In this work we applied metaheuristics for determining the optimal hyperparameters for each case. Using automatic gradient calculation we can differentiate the neural network by its input variables and construct a physics error function $f$. The loss function involves a loss term for the data and a loss term for the physics function.}
\label{fig:pinn}
\end{figure}

Any neural network such as fully connected, recurrent, and convolutional networks can be used, where we make use of automatic gradient calculations as used by back-propagation to compute derivatives of the output variables with respect to the input variables. An example of such a network is shown in Fig.~\ref{fig:pinn}, with $f$ representing the PDE rewritten as $f = 0$. 

The network is designed with input variables $x$, $y$, and $t$ and output variables  $\hat{u}$ and $\hat{v}$. The output variables are used directly to calculate the data loss term, while for the physics loss term we differentiate the variables with respect to the input variables as needed for the physics loss function. Observe that the neural network must have input variables that correspond to the physics law's derivative terms. That is, if $\nicefrac{\partial u}{\partial x}$ is part of our physics loss function, then $x$ must be an input variable and $u$ an output variable of our neural network.

The objective is to train both in a supervisory manner with measured or simulated data (i.\,e.~$\ell_\text{data}$), as well as training to minimize the departure from the physics law. The $\ell_\text{physics}$ term ensures that the neural network generalizes better for unseen data by preventing overfitting. The physics loss encourages that the output variables are not just trained to a local region around the input values of the given data, but that also their first and/or second-degree derivations (depending on the PDEs) match with our physical understanding of the model thus spurring better predictive power outside the region of the training data.

We denote $\mathbb{N}_l \times \mathbb{M}_l$ as a multi-layer perceptron (MLP) of $\mathbb{N}_l$ hidden layers and $\mathbb{M}_l$ neurons per layer. As part of the experimentation we assess different activation functions. As our optimization problem is non-convex, the stochastic gradient-descent Adam optimizer was used~\cite{kingma2017Adam} with different learning rates depending on the problem. The weights of the neural network are initialized randomly using Xavier's initialization method~\cite{glorot2010xavier}. As we intend to analyze how activation functions play a role in learning the derivations of PDEs the $\tanh$, GELU, Softplus, LogSigmoid, Sigmoid, TanhShrink, CELU, Softsign, and ReLU activation functions.% \luis{cita}

We select a sample subset $(X_i, Y_i) \forall i \in [0,N_u-1]$ at random from the entire solution space for training the data loss, with $X_i \in \mathbb{R}^D$ the $D$ features and $Y_i \in \mathbb{R}^P$ the labels of dimension $P$. For the physics loss we select a subset $X_i \forall i \in [0,N_f-1]$ at random for which we evaluate the PDE error. Note that for the physics loss the solution is not needed, allowing applications to be able to train even when few data are available of the solution. In general we pick $N_u \ll N_f$.

\section{Experiments}\label{sec:experiments}

As stated in Section \ref{sec:foundations}, we consider three models: the first ones, the Burgers equation, is a nonlinear one-dimensional and the other two, wave, and advection-diffusion equations, are linear and two dimensional PDEs.

%( as stated in...) Each of these equations is defined in $\Omega \times \mathbb{R}_0^+$, where $\Omega \subseteq \mathbb{R}^n$ is an open set, and we consider $n=2$.

\subsection{Datasets preparation}

In order to generate data for training, validation, and testing, we simulate the PDEs using a Fourier spectral method for the one dimensional model, finite element methods (FEM) for the wave equation and the explicit analytical solution for the advection-diffusion equation. Simulations for the two-dimensional problems were performed using FEniCS \citep{AlnaesBlechta2015a}. %The spatial domain, $\Omega = \left]0,L_x\right[\times\left]0,L_y\right[$ with $L_x = 1$, $L_y =1.5$, is represented by an unstructured rectangular mesh with 218 nodes and 384 triangles. For the time scheme, we take the domain $[0, T_f]$ with $n$ uniform distributed time steps, so the discretized domain is $\{ 0, dt, 2 dt,\dots, n dt \}$, where $dt = \frac{T_f}{n}$. Now we give more details of the chosen values in each model.

\paragraph{Burgers equation}
We consider the following Burgers equation
\begin{equation}
\frac{\partial u}{\partial t} + u \frac{\partial u}{\partial x} = \nu \frac{\partial^2 u}{\partial x^2}\,,
\end{equation}
where $\nu$ is a diffusion coefficient. This equation usually appears in the context of fluid mechanics, and more specifically models one-dimensional internal waves in deep ocean. It represents a hyperbolic conservation law as $\nu \rightarrow 0$ and  it is the simplest model for analyzing the effect of nonlinear advection and diffusion in a combined way. Notice that the Burgers equation can be written as a physics loss term of \eqref{eqn:full-loss} as  
%\begin{equation}
$f_\text{Burgers} [u] = 0.$ %\,.
%\label{eq:burgers}
%\end{equation}
%\hugo{es $f_{burgers}$, ya que me refiero a la funcion $f$ que está en la intro de PINNs}
%\luis{lo he puesto inline para ahorrar espacio.}
%\hugo{ok}

We simulated the Burgers equation in the spatiotemporal domain $[0, L] \times [0,T]$, using a Fourier spectral method as described in \cite{basdevant1986spectral}. % which is implemented in MATLAB. 

The diffusion coefficient is taken as $\displaystyle \nu = \left( \frac{0.01}{\pi} \right)$. 

We consider the initial condition 
\begin{equation}
u(x,0) = -\sin(\pi x),\ \text{with}\ x \in \left(-1,1\right)\,,
\end{equation}
and boundary conditions
\begin{equation}
    u(1,t) = u(-1,t) = 0,\ \text{with}\  t \in (0,1)\,,
\end{equation}
with $N=512$ spatial points in a uniform grid in $\left[-1, 1\right]$ and $100$ points in time defined in a uniform grid in $[0,1]$.  

\paragraph{Wave equation}

In this case, we consider the wave equation% in $\Omega \times \mathbb{R}_0^+$, where $\Omega \subseteq \mathbb{R}^n$ is an open set: % in two dimensions with initial and boundary conditions:
\begin{equation}
\frac{\partial^2 \eta }{\partial t^2} - \nabla \cdot \left( H \nabla \eta \right)  = 0\,, %& \text{in}\ \Omega 
%\begin{array}{rll}
%\displaystyle \frac{\partial^2 \eta }{\partial t^2} - \nabla \cdot \left( H \nabla \eta \right) & = 0 %& \text{in}\ \Omega \times ]0,T[\,,\\
%\eta(x,y,t) & = \eta_{\partial} (x,y,t) & \text{on}\ \partial \Omega \times ]0,T[\,, \\
%\multicolumn{2}{c}{\eta(x,y,t) = \eta_0 (x,y)\ \text{and}\ \frac{\partial \eta}{\partial t} (x,y,0) = \eta_1} & \text{in}\ \Omega\,.
%\end{array}
\end{equation}
where $\eta: \overline{\Omega} \times \mathbb{R}_0^+ \rightarrow \mathbb{R}$ represents the superficial fluctuations of a water container or channel and $H:\overline{\Omega} \rightarrow \mathbb{R}_0^+$ is the depth of the water from a reference level. 
%This equation can be obtained from the shallow water equations under certain conditions (CITE!). 
This equation can be written as
%\begin{equation}
$f_\text{waves} [\eta, H] = 0$. %\,.
%\label{eq:wave}
%\end{equation}
This is where the physics loss is taken from for this PDE. 

We simulate the wave equation in a rectangular domain in the time interval $]0,T_f[$, with Dirichlet boundary conditions for $\eta$ as
\begin{equation}
\eta = 0, \ \text{on} \ \partial \Omega \times \left]0,T_f\right[\,,
\end{equation}
and initial conditions on $\Omega$
\begin{equation}
\begin{array}{rl}
\eta (x,y,0) & = \exp \left( -10 \cdot \left((x-0.5)^2 + (y - 0.75)^2 \right) \right)\,,\\
\frac{\partial \eta}{\partial t} (x,y,0) & = 0\,.
\end{array}
\end{equation}
On the other hand, the depth $H$ is given by
\begin{equation}
H(x,y) = (1-x)(2-\sin(3\pi y)) \quad \text{in} \ \overline{\Omega}\,.
\end{equation}
We implement FEM in this equation considering Lagrange finite elements of degree 1. The time scheme is explicit and $T_f = 1.0$, $n=100$.

\paragraph{Advection-diffusion equation}

We consider the advection-diffusion equation for heat transfer, assuming that the diffusion coefficient is constant, there are no sources nor sinks of heat, and the temperature depends only on $x$, $y$, and $t$, that is, 
\begin{equation}
\frac{\partial T}{\partial t} = D \nabla^2 T - \boldsymbol{u} \cdot \nabla T\,,
\end{equation}
where $T: \overline{\Omega} \times \mathbb{R}_0^+ \rightarrow \mathbb{R}$ is the temperature in $K$, $D = 0.1 \frac{\text{m}^2}{\text{s}}$ is the thermal diffusivity considered in this work, and $\boldsymbol{u}:\overline{\Omega} \times \mathbb{R}_0^+ \rightarrow \mathbb{R}^2$ the velocity field in $\frac{m}{s}$. The equation can be written as a physics loss term as
\begin{equation}
f_{AD} [u,v,T] = 0\,.
\end{equation}

We simulate the advection-diffusion equation in $\Omega \times (0,T_f)$ where $\Omega = (0,L_x) \times (0, L_y)$, where $L_x = L_y = 1.0$ and $T_f = 1.0$. In addition, $D = 0.02$, $u = (\cos(\phi)$ and $v = \sin(\phi))$, where $\phi = 22.5^\circ$. %As mentioned in \cite{he2020physicsinformed}, 
We consider the following initial conditions
\begin{equation}
    T(x,y,0) = \exp\left( -\frac{x^2 + y^2}{D} \right), \ \text{in} \Omega\,,
\end{equation}
and boundary condition
\begin{equation}
        T(x,y, t) = \frac{1}{4t + 1} \exp\left[ - \frac{x^2 + y^2 + t^2 (u^2 + v^2) - 2t(xu + yv) }{D (4t+1)} \right] \qquad \text{on } \partial \Omega\,.
\end{equation}
As the coefficients are constant and the domain is a square, we are able to solve this linear equation by known analytic methods. The explicit solution of this equation is 
\begin{equation}
    T(x,y, t) = \frac{1}{4t + 1} \exp\left[ - \frac{x^2 + y^2 + t^2 (u^2 + v^2) - 2t(xu + yv) }{D (4t+1)} \right]\,, 
\end{equation}
so we generate this solution in a uniform unit square mesh with 40 points in each spatial dimension. 

%5the same spatial and temporal domain as the shallow water equations \taco{shallow water section is removed, put temporal domain here} but in this case we consider mixed Dirichlet and Neumann boundary conditions\luis{cita} for the temperature (in ºC):
%\begin{equation}
%T = 0 \quad \text{on} \ \{0,L_x\}\times [0,L_y], \qquad \frac{\partial T}{\partial n} = 0 \quad \text{on} \ [0,L_x] \times \{0,L_y\}\,, 
%\end{equation}
%and initial condition (in ºC):
%\begin{equation}
%T(x,y,0) = 15 + 45 \cdot \exp\left[ -(x-0.5)^2 - (y-0.75)^2\right]\,.
%\end{equation}
%We performed a $\theta$-scheme in time with $\theta = 0.75$, and $T_f = 1$, $n=500$. The velocity is given by the solution of the shallow water system.\taco{no, we fix the velocity field}

\section{Results and discussion}\label{sec:results-discussion}

%- traer aquí section (Discussion)

\begin{figure}[t]
  \centering
  \includegraphics[width=\textwidth]{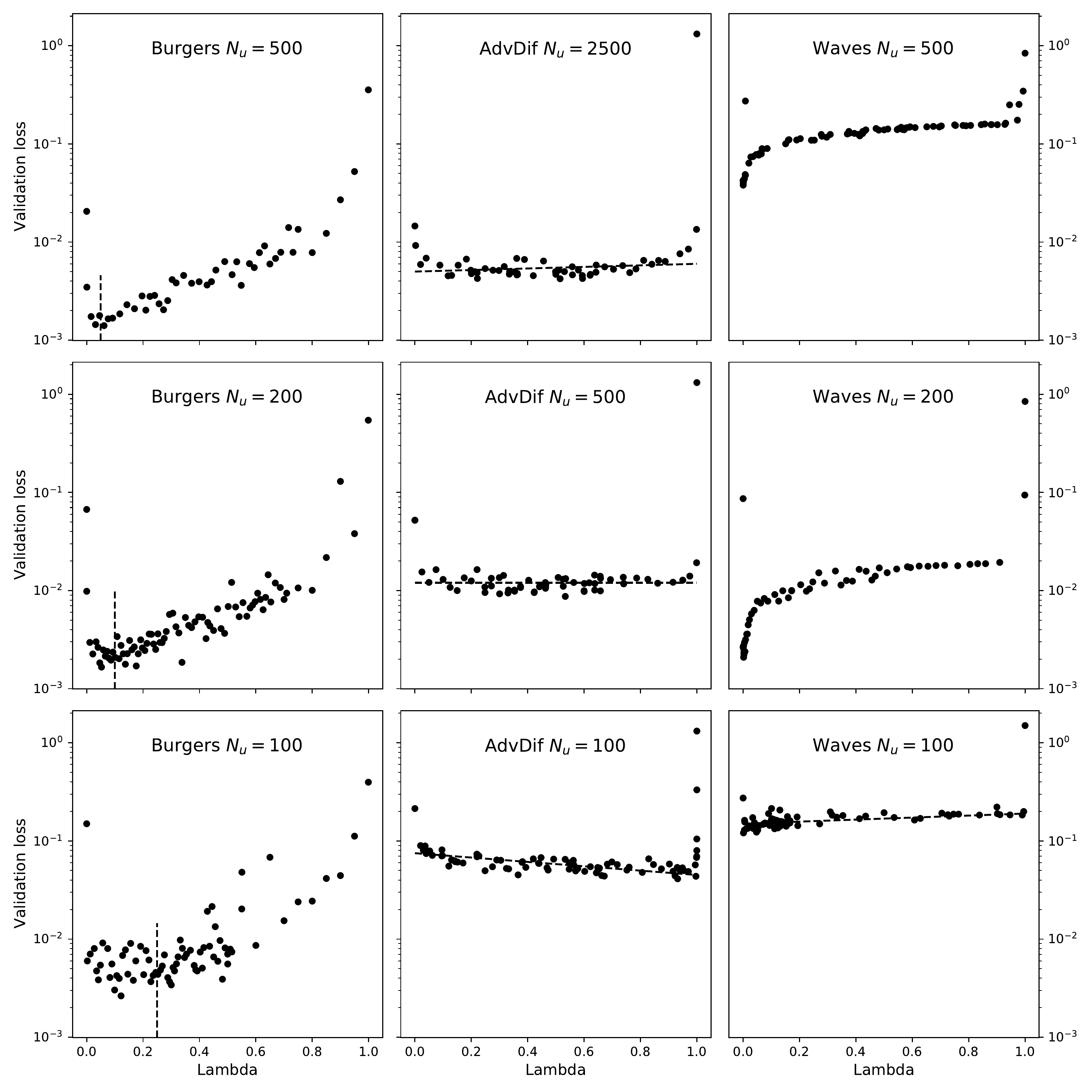}
  \caption{The relative error for different values of $N_u$ and for different problems. Each dot represents the final value of the relative error after training. We observe how the choice of $\lambda$ affects training results, and we observe a slight shift to higher $\lambda$ values as $N_u$ decreases. The dashed lines were added manually to aid in observing trends. For Burgers we note that the minimum shifts towards $1.0$ as $N_u$ decreases, and for advection-diffusion the slope becomes more negative as $N_u$ decreases. The wave equation is more less evident, but it can be seen that for $N_u=500$ and $N_u=200$ the minimum lies around zero, while for $N_u=100$ the trend has flattened and the results would be similar for $\lambda \in \left[0.1,0.9\right]$ roughly.}
  \label{fig:lambda}
\end{figure}

\begin{table}[tb]
    \centering
    \caption{Parameters for each of the models. Here epochs are the numbers of iterations over the training set, and $N_u$ and $N_f$ are the number of points used for the data loss and the physics loss respectively. $N_u^{val}$ is the number of data points used to for the validation data loss.}
    \label{table:model_params}
    \medskip
    \begin{tabular}{@{}rcccccc@{}}
        \toprule
        \textbf{Model} & \textbf{NN setup} & \textbf{Optimizer} & \textbf{Epochs} & $N_u$ & $N_f$ & $N_u^{val}$ \\
        \midrule
        Burgers & $8 \times 20$ & Adam (lr=0.001) & 25000 & 100, 200, 500 & 12800 & 25600 \\
        Advec.-Diffusion & $3 \times 40$ & Adam (lr=0.002) & 20000 & 100, 500, 2500 & 5000 & 168100 \\
        Wave & $5 \times 50$ & Adam (lr=0.001) & 10000 & 100, 200, 500 & 10000 & 21800 \\
        %ShaWat & $5 \times 100$ & Adam lr=0.001 & 10000 & 200 & 10000 & ? \\
        \bottomrule \\ 
    \end{tabular}
\end{table}

See Table~\ref{table:model_params} for an overview of the used parameters for each of the models. The hyperparameters, such as the number and size of the hidden layers; the optimizer; and the number of epochs, were chosen either manually or by a grid search. All experiments were executed using Google Colab, a platform to run notebooks on fast graphics processing units (GPUs). The ability of GPUs to run calculations in parallel greatly enhances the speed of evaluating the neural network and calculating the gradients.\footnote{The data sets, source code, and results are available online at \url{https://github.com/Inria-Chile/assessing-pinns-ocean-modelling} and are released under the CeCILL license.}

To be able to compare the performance between models, the validation data loss is calculated over the entire data set and then normalized as
\begin{equation}
    \textrm{Relative Error} = \frac{\sqrt{\textrm{MSE}(Y_{val},\hat{Y}_{val})}}{\left\lVert Y_{val}\right\rVert_2}\,, %\sqrt{\nicefrac{\textrm{MSE}(Y_{val},\hat{Y}_{val})}{\left\lVert Y_{val}\right\rVert_2^2}}
\end{equation}
where $Y_{val}$ are the labels of the validation data set, $\hat{Y}_{val}$ the output of the neural network, and $\textrm{MSE}$ the mean squared error. In the case of advection-diffusion the labels $Y_i$ have been normalized to have zero mean and unit standard deviation to bring down the scale of the temperature, which is in Kelvin, to a range around zero. This obviates the need for the division when calculating the relative error. Additionally, to reduce the jitter in the loss while training, we use the lowest validation data loss of the last 250 epochs to calculate the relative error.

Learning the optimal value of $\lambda$, the relative weighting of the physics loss with respect to the data loss is a metalearning objective in order to improve subsequent training of the same PINN. In Fig.~\ref{fig:lambda} we show the results of evaluating different problems using different values for $N_u$ to find an optimal value of $\lambda$. A sharp rise in the relative error can be observed at $\lambda=0.0$ and $\lambda=1.0$, which are equivalent to training using solely the data loss or physics loss terms respectively. We note that the optimal value for $\lambda$ depends on the problem at hand and specifically on the scale of the data loss versus the physics loss. Normalizing the data set labels would in part solve the problem, but the scale of the physics loss remains hard to normalize. In our experiments however, we note that for typical values of $\lambda$ the comparison of the data and physics loss remains roughly $0.1 < \ell_\text{data}/\ell_\text{physics} < 10$. %This suggests that optimal values of $\lambda$ should be shall not be at the extremes, say, within $1\cdot10^{-6}$ of $\lambda=0.0$ or $\lambda=1.0$.

Per problem we observe that at decreasing values for $N_u$, the optimal value of $\lambda$ shifts slightly towards $1.0$. This trend suggests that as data become more sparse, the physics loss takes higher importance. As less data are available, the information input from the physics loss helps training significantly, while the reverse holds as more data are available. At the limit where $N_u$ is sufficiently large to be able to train the network autonomously, the optimal value of $\lambda$ tends to zero.

\begin{figure}[t]
\centering
\begin{subfigure}{.333\textwidth}
  \centering
  \includegraphics[width=\linewidth]{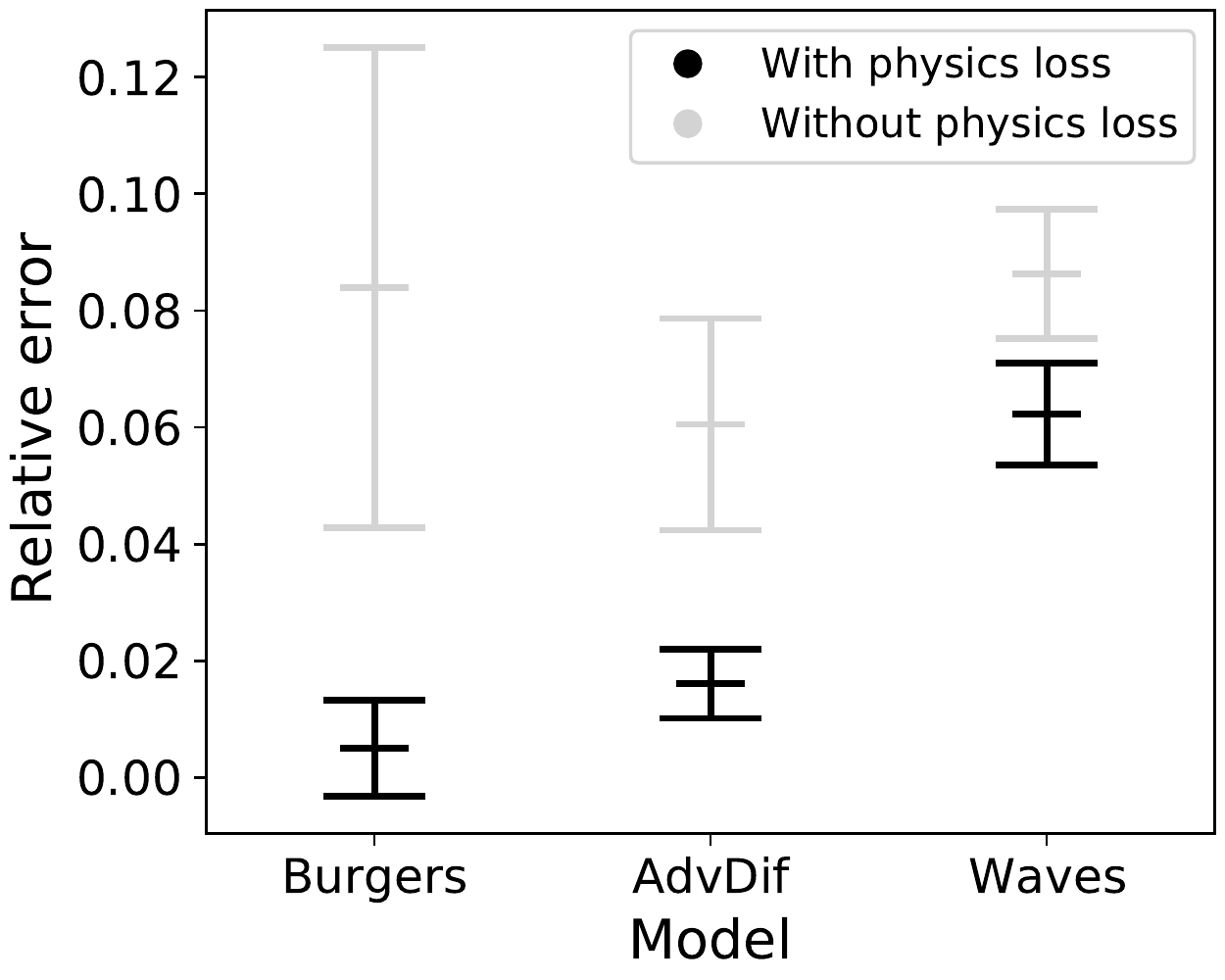}
  \caption{Errors with/without physics loss.}
  \label{fig:robustness}
\end{subfigure}\hfill%
\begin{subfigure}{.667\textwidth}
  \centering
  \includegraphics[width=\linewidth]{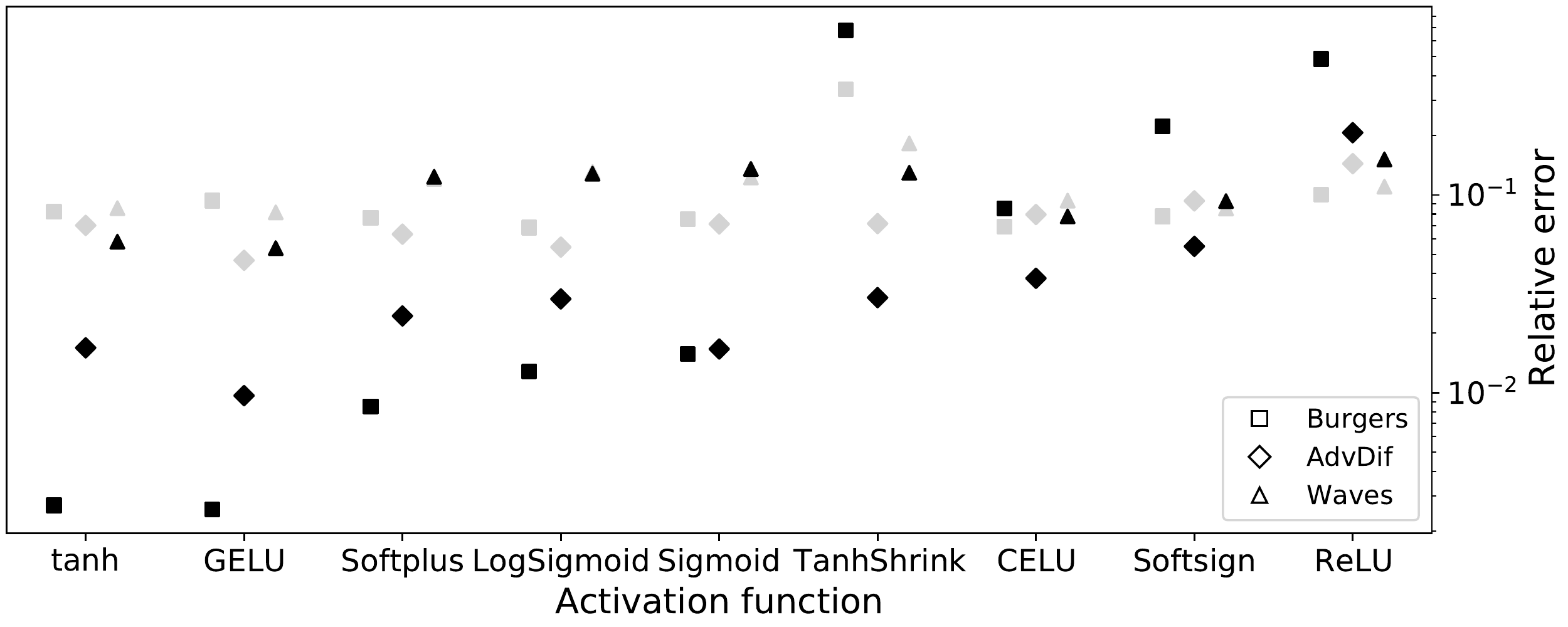}
  \caption{Relative error using different activation functions.}
  \label{fig:funcs}
\end{subfigure}
\caption{Summary of the experimental results. Fig. \ref{fig:robustness} shows results for 25 runs with and without physics loss for each of the models. For Fig.~\ref{fig:funcs}, $N_u$ is $200$, $500$, and $200$ and $\lambda$ is $0.1$, $0.5$, and $0.002$ for the Burgers, advection-diffusion, and wave equations respectively.}
\label{fig:loss}
\end{figure}

In cases where few data points are known (i.e. $N_u$ is small), the choice of the sample subset can have a large influence on training results. In Fig.~\ref{fig:robustness} the relative error is shown for 25 random data subset selections with and without the physics loss. The added physics loss term reduces the relative error, averaged over all three models, by $65\%$ and reduces its standard deviation by $56\%$. PINNs thus improve training results and reduce variability when the number of data points is low.

An analysis of how activation functions play a role in learning the derivations of PDEs, a comparison is made with nine activation functions in Fig.~\ref{fig:funcs}. Here we compare the $\tanh$, GELU, Softplus, LogSigmoid, Sigmoid, TanhShrink, CELU, Softsign, ReLU activation function for their performance. It can be observed that especially the $\tanh$ and GELU activation functions and to a lesser extend Softplus work well. The ReLU function produces poor results due to its inability to learn second or higher derivatives. Due to the fact that a DNN using the ReLU activation function produces a piece-wise linear function~\cite{liu2021relu}, its second derivative is zero except at the points where the linear segments meet, where it will be a Dirac function at $x=0$, inhibiting learning second derivatives or higher. The Softsign and CELU functions have similar problems given their discontinuity of the second derivative at $x=0$.

\section{Final remarks}\label{sec:final}

Information from the PDE serves in cases where data availability is limited to the extent that it would either be unable to train the network satisfactorily using solely the data, or where results would vary widely depending on the selection of the data points. The addition of the physics loss term both improves training results and improves the stability of the training, even more so when the weight in the loss function is optimal.

Finding the optimal weight of the loss function is a complex problem and this is, to the best of our knowledge, the first study to understand the behavior of it with respect to different hyperparameters and PDEs. In the future, we expect to find the optimal weights using only the data available. In addition, it is also the first attempt at the study of PINNs for ocean modelling equations with very simplified geometries. We expect to extend our results to more complex and realistic geometries in order to use our results in real applications.

As $\lambda$ has a dual role: the prior confidence in the training data vs. the PDE and the scaling of these two terms, future work could understand each role of this parameter, which could be a possible explanation of why the optimal weight in some of our experiments are so different between different models. In addition, as we observed for very small data sets, it would be interesting to find what ``good'' sample subsets of the problem for training are more stable under optimization. This would be the first step to estimate the minimal amount of data required to train the neural networks in a robust way. 

As the equations of interest are inherently wave-like, we propose to explore the benefits of Fourier transformations for learning the equations more effectively, as was done by~\cite{li2020fourier}. This would enable the network to learn patterns of the data in the frequency space. Additionally, to generalize the model better for unseen data, we propose the use of dropout by~\cite{Hinton2012} to randomly turn off a percentage of the neurons. This forces the network to learn redundancies in case certain neurons are turned off, increasing the stability of the network, and the likelihood to generalize better for unseen data.

Future work on this topic is needed as it is required to develop better models for understanding the ocean and integrate those models with others that explain the interaction between organisms and the environment. This is essential because of the constantly changing nature of the ocean.

We believe that PINNs have the potential of providing such tools, and therefore, serve as a building block of a more comprehensive solution that addresses the ultimate issue of understanding and mitigating the climate change crisis. However, this potential also comes with important risks as using an inadequate, biased, or non-transparent tool could lead instead to a deterioration. In this direction, as we already mentioned in the paper, we are interested in combining PINNs with explainable AI and powerful visualization tools. Similarly, extended validation is necessary by domain experts.

\begin{ack}
This work is funded by project CORFO 10CEII-9157 Inria Chile and Inria Challenge project OcéanIA (desc. num 14500).
\end{ack}

\bibliography{nn-solver, oceania}

\begin{thebibliography}{}

\bibitem[Aln{\ae}s et~al., 2015]{AlnaesBlechta2015a}
Aln{\ae}s, M.~S., Blechta, J., Hake, J., Johansson, A., Kehlet, B., Logg, A.,
  Richardson, C., Ring, J., Rognes, M.~E., and Wells, G.~N. (2015).
\newblock The {FEniCS} project version 1.5.
\newblock {\em Archive of Numerical Software}, 3(100).

\bibitem[Basdevant et~al., 1986]{basdevant1986spectral}
Basdevant, C., Deville, M., Haldenwang, P., Lacroix, J., Ouazzani, J., Peyret,
  R., Orlandi, P., and Patera, A. (1986).
\newblock Spectral and finite difference solutions of the burgers equation.
\newblock {\em Computers \& fluids}, 14(1):23--41.

\bibitem[Burgers, 1948]{BURGERS1948171}
Burgers, J. (1948).
\newblock A mathematical model illustrating the theory of turbulence.
\newblock In {Von Mises}, R. and {Von Kármán}, T., editors, {\em Advances in
  Applied Mechanics}, volume~1, pages 171--199. Elsevier.

\bibitem[Chen et~al., 2018]{chen-2018:neural-odes}
Chen, R.~T., Rubanova, Y., Bettencourt, J., and Duvenaud, D.~K. (2018).
\newblock Neural ordinary differential equations.
\newblock In {\em Advances in neural information processing systems}, pages
  6571--6583.

\bibitem[de~Wolff et~al., 2021]{tacohugo-2021:aimocc}
de~Wolff, T., Carrillo~Lincopi, H., Mart\'i, L., and Sanchez-Pi, N. (2021).
\newblock Assessing physics informed neural networks in ocean modelling and
  climate change applications.
\newblock In Sanchez-Pi, N. and Mart\'i, L., editors, {\em AI: Modeling Oceans
  and Climate Change Workshop at ICLR 2021}.

\bibitem[Dupont et~al., 2019]{dupont-2019:augmented-odes}
Dupont, E., Doucet, A., and Teh, Y.~W. (2019).
\newblock Augmented neural {ODE}s.
\newblock In Wallach, H., Larochelle, H., Beygelzimer, A., d\textquotesingle
  Alch\'{e}-Buc, F., Fox, E., and Garnett, R., editors, {\em Advances in Neural
  Information Processing Systems 32}, pages 3140--3150. Curran Associates, Inc.

\bibitem[Glorot and Bengio, 2010]{glorot2010xavier}
Glorot, X. and Bengio, Y. (2010).
\newblock Understanding the difficulty of training deep feedforward neural
  networks.
\newblock In Teh, Y.~W. and Titterington, M., editors, {\em Proceedings of the
  Thirteenth International Conference on Artificial Intelligence and
  Statistics}, volume~9 of {\em Proceedings of Machine Learning Research},
  pages 249--256, Chia Laguna Resort, Sardinia, Italy. PMLR.

\bibitem[He and Tartakovsky, 2020]{he2020physics}
He, Q. and Tartakovsky, A.~M. (2020).
\newblock Physics-informed neural network method for forward and backward
  advection-dispersion equations.
\newblock {\em arXiv preprint arXiv:2012.11658}.

\bibitem[Hinton et~al., 2012]{Hinton2012}
Hinton, G.~E., Srivastava, N., Krizhevsky, A., Sutskever, I., and
  Salakhutdinov, R. (2012).
\newblock Improving neural networks by preventing co-adaptation of feature
  detectors.
\newblock {\em CoRR}, abs/1207.0580.

\bibitem[Kingma and Ba, 2017]{kingma2017Adam}
Kingma, D.~P. and Ba, J. (2017).
\newblock {ADAM}: {A} method for stochastic optimization.

\bibitem[Li et~al., 2020]{li2020fourier}
Li, Z., Kovachki, N., Azizzadenesheli, K., Liu, B., Bhattacharya, K., Stuart,
  A., and Anandkumar, A. (2020).
\newblock Fourier neural operator for parametric partial differential
  equations.

\bibitem[Liu and Liang, 2021]{liu2021relu}
Liu, B. and Liang, Y. (2021).
\newblock Optimal function approximation with {ReLU} neural networks.
\newblock {\em Neurocomputing}, 435:216--227.

\bibitem[L\"utjens et~al., 2021]{bjorn-2021:aimocc}
L\"utjens, B., Crawford, C.~H., Veillette, M., and Newman, D. (2021).
\newblock {PCE-PINNs}: {P}hysics-informed neural networks for uncertainty
  propagation in ocean modeling.
\newblock In Sanchez-Pi, N. and Mart\'i, L., editors, {\em AI: Modeling Oceans
  and Climate Change Workshop at ICLR 2021}.

\bibitem[McLean, 2012]{mclean2012understanding}
McLean, D. (2012).
\newblock {\em Understanding aerodynamics: arguing from the real physics}.
\newblock John Wiley \& Sons.

\bibitem[Orlandi, 2000]{Orlandi2000}
Orlandi, P. (2000).
\newblock {\em The Burgers equation}, pages 40--50.
\newblock Springer Netherlands, Dordrecht.

\bibitem[Raissi et~al., 2019]{RAISSI2019686}
Raissi, M., Perdikaris, P., and Karniadakis, G. (2019).
\newblock Physics-informed neural networks: A deep learning framework for
  solving forward and inverse problems involving nonlinear partial differential
  equations.
\newblock {\em Journal of Computational Physics}, 378:686--707.

\bibitem[Rao, 2010]{rao2010introduction}
Rao, K.~S. (2010).
\newblock {\em Introduction to partial differential equations}.
\newblock PHI Learning Pvt. Ltd.

\bibitem[S\'{a}nchez-Pi et~al., 2020]{sanchez-pi-et-al-2020--neurips}
S\'{a}nchez-Pi, N., Mart\'{i}, L., Abreu, A., Bernard, O., de~Vargas, C.,
  Eveillard, D., Maass, A., Marquet, P.~A., Sainte-Marie, J., Salomon, J.,
  Schoenauer, M., and Sebag, M. (2020).
\newblock Artificial intelligence, machine learning and modeling for
  understanding the oceans and climate change.
\newblock In Dao, D., Sherwin, E., Donti, P., Kaack, L., Kuntz, L., Yusuf, Y.,
  Rolnick, D., Nakalembe, C., Monteleoni, C., and Bengio, Y., editors, {\em
  Tackling Climate Change with Machine Learning workshop at NeurIPS 2020}.

\bibitem[Sirignano and Spiliopoulos, 2018]{Sirignano}
Sirignano, J. and Spiliopoulos, K. (2018).
\newblock {DGM}: {A} deep learning algorithm for solving partial differential
  equations.
\newblock {\em Journal of Computational Physics}, 375:339--1364.

\bibitem[Stoker, 2011]{stoker2011water}
Stoker, J.~J. (2011).
\newblock {\em Water waves: The mathematical theory with applications},
  volume~36.
\newblock John Wiley \& Sons.

\bibitem[Temam, 2001]{temam2001navier}
Temam, R. (2001).
\newblock {\em {N}avier-{S}tokes equations: {T}heory and numerical analysis},
  volume 343.
\newblock American Mathematical Soc.

\bibitem[Union, 2013]{union2013advection}
Union, J. I.~G. (2013).
\newblock Advection diffusion equation models in near-surface geophysical and
  environmental sciences.
\newblock {\em J. Ind. Geophys. Union (April 2013)}, 17(2):117--127.

\bibitem[van~der Meer et~al., 2020]{van2020optimally}
van~der Meer, R., Oosterlee, C., and Borovykh, A. (2020).
\newblock Optimally weighted loss functions for solving {PDEs} with neural
  networks.
\newblock {\em arXiv preprint arXiv:2002.06269}.

\bibitem[van Herten et~al., 2020]{van2020physics}
van Herten, R.~L., Chiribiri, A., Breeuwer, M., Veta, M., and Scannell, C.~M.
  (2020).
\newblock Physics-informed neural networks for myocardial perfusion mri
  quantification.
\newblock {\em arXiv preprint arXiv:2011.12844}.

\bibitem[Vetra-Carvalho et~al., 2018]{doi:10.1080/16000870.2018.1445364}
Vetra-Carvalho, S., van Leeuwen, P.~J., Nerger, L., Barth, A., Altaf, M.~U.,
  Brasseur, P., Kirchgessner, P., and Beckers, J.-M. (2018).
\newblock State-of-the-art stochastic data assimilation methods for
  high-dimensional non-{G}aussian problems.
\newblock {\em Tellus A: Dynamic Meteorology and Oceanography}, 70(1):1--43.

\bibitem[Xiang et~al., 2021]{xiang2021self}
Xiang, Z., Peng, W., Zheng, X., Zhao, X., and Yao, W. (2021).
\newblock Self-adaptive loss balanced physics-informed neural networks for the
  incompressible navier-stokes equations.
\newblock {\em arXiv preprint arXiv:2104.06217}.

\end{thebibliography}
\bibliographystyle{apalike}

\section*{Checklist}

% %%% BEGIN INSTRUCTIONS %%%
% The checklist follows the references.  Please
% read the checklist guidelines carefully for information on how to answer these
% questions.  For each question, change the default \answerTODO{} to \answerYes{},
% \answerNo{}, or \answerNA{}.  You are strongly encouraged to include a {\bf
% justification to your answer}, either by referencing the appropriate section of
% your paper or providing a brief inline description.  For example:
% \begin{itemize}
%   \item Did you include the license to the code and datasets? \answerYes{See Section~\ref{sec:results-discussion}.}
%   \item Did you include the license to the code and datasets? \answerNo{}
%   \item Did you include the license to the code and datasets? \answerNA{}
% \end{itemize}
% Please do not modify the questions and only use the provided macros for your
% answers.  Note that the Checklist section does not count towards the page
% limit.  In your paper, please delete this instructions block and only keep the
% Checklist section heading above along with the questions/answers below.
% %%% END INSTRUCTIONS %%%

\begin{enumerate}

\item For all authors...
\begin{enumerate}
  \item Do the main claims made in the abstract and introduction accurately reflect the paper's contributions and scope?
    \answerYes{}
  \item Did you describe the limitations of your work?
    \answerYes{}
  \item Did you discuss any potential negative societal impacts of your work?
    \answerYes{See in particular Section \ref{sec:final}}
  \item Have you read the ethics review guidelines and ensured that your paper conforms to them?
    \answerYes{}
\end{enumerate}

\item If you are including theoretical results...
\begin{enumerate}
  \item Did you state the full set of assumptions of all theoretical results?
    \answerYes{}
	\item Did you include complete proofs of all theoretical results?
    \answerNA{}
\end{enumerate}

\item If you ran experiments...
\begin{enumerate}
  \item Did you include the code, data, and instructions needed to reproduce the main experimental results (either in the supplemental material or as a URL)?
    \answerYes{See Section~\ref{sec:results-discussion}.}
  \item Did you specify all the training details (e.g., data splits, hyperparameters, how they were chosen)?
    \answerYes{See Section~\ref{sec:results-discussion}.}
	\item Did you report error bars (e.g., with respect to the random seed after running experiments multiple times)?
    \answerYes{See Section~\ref{sec:results-discussion}.}
	\item Did you include the total amount of compute and the type of resources used (e.g., type of GPUs, internal cluster, or cloud provider)?
    \answerYes{See Section~\ref{sec:results-discussion}.}
\end{enumerate}

\item If you are using existing assets (e.g., code, data, models) or curating/releasing new assets...
\begin{enumerate}
  \item If your work uses existing assets, did you cite the creators?
    \answerNA{}
  \item Did you mention the license of the assets?
    \answerYes{See Section~\ref{sec:results-discussion}.}
  \item Did you include any new assets either in the supplemental material or as a URL?
    \answerYes{See Section~\ref{sec:results-discussion}.}
  \item Did you discuss whether and how consent was obtained from people whose data you're using/curating?
    \answerNA{}
  \item Did you discuss whether the data you are using/curating contains personally identifiable information or offensive content?
    \answerNA{}
\end{enumerate}

\item If you used crowdsourcing or conducted research with human subjects...
\begin{enumerate}
  \item Did you include the full text of instructions given to participants and screenshots, if applicable?
    \answerNA{}
  \item Did you describe any potential participant risks, with links to Institutional Review Board (IRB) approvals, if applicable?
    \answerNA{}
  \item Did you include the estimated hourly wage paid to participants and the total amount spent on participant compensation?
    \answerNA{}
\end{enumerate}

\end{enumerate}

%%%%%%%%%%%%%%%%%%%%%%%%%%%%%%%%%%%%%%%%%%%%%%%%%%%%%%%%%%%%

%\appendix
%\section{Appendix}
%Optionally include extra information (complete proofs, additional experiments and plots) in the appendix.
%This section will often be part of the supplemental material.

\end{document}